\documentclass[1p,review]{elsarticle}

\usepackage{amsmath,amsfonts}
\usepackage{algorithm}
\usepackage{algorithmic}
\usepackage{array}
\usepackage{booktabs}
\usepackage{multirow}
\usepackage{xcolor}
\usepackage{colortbl}
\usepackage{subfigure}
\usepackage{textcomp}
\usepackage{float}
\usepackage{stfloats}
\usepackage{url}
\usepackage{verbatim}
\usepackage{graphicx}
\usepackage[colorlinks=true, citecolor=blue, linkcolor=blue, urlcolor=blue]{hyperref}
\usepackage[capitalise, noabbrev]{cleveref}
\graphicspath{{./figures/}}

\journal{Engineering Applications of Artificial Intelligence}

\begin{document}
\begin{frontmatter}

\title{An Efficient Additive Kolmogorov-Arnold Transformer for Point-Level Maize Localization in Unmanned Aerial Vehicle Imagery}

\author[uwbse]{Fei Li}
\ead{leefly072@126.com}
\author[uwbse]{Lang Qiao}
\ead{lqiao@umn.edu}
\author[uwbse]{Jiahao Fan}
\ead{jiahao.fan@wisc.edu}
\author[uwbse]{Yijia Xu}
\ead{xu556@wisc.edu}
\author[agr]{Shawn M. Kaeppler}
\ead{smkaeppl@wisc.edu}
\author[uwbse]{Zhou Zhang\corref{cor1}}
\ead{zzhang347@wisc.edu}

\cortext[cor1]{Corresponding author}

\address[uwbse]{Department of Biological Systems Engineering, University of Wisconsin-Madison, Madison, WI 53706, USA}
\address[agr]{Department of Agronomy, University of Wisconsin-Madison, Madison, WI 53706, USA}


\begin{abstract}
Accurate identification of maize plants at the point level from unmanned aerial vehicle imagery is a key requirement for high-throughput field phenotyping and automated stand assessment in precision agriculture. This study presents the Additive Kolmogorov--Arnold Transformer, an efficient framework designed to enhance the representation of fine-grained plant structures and to support reliable analysis of large-scale aerial imagery. The model employs Pad{\'e} Kolmogorov--Arnold Network modules in place of conventional multilayer perceptrons, together with a novel additive attention mechanism that captures multiscale spatial patterns while reducing computational cost. To facilitate benchmarking under realistic field conditions, we also introduce the Point-based Maize Localization dataset, consisting of 1{,}928 high-resolution images with approximately 501{,}000 manually annotated plant locations. Experimental results show that the proposed approach achieves a mean of precision and recall of 62.8\%, outperforming existing methods and offering improvements in both computational efficiency and throughput. In downstream agricultural applications, the model provides accurate stand counts, with a mean absolute error of 7.1, and precise inter-plant spacing estimates, with a root mean square error of 1.95--1.97~cm. These outcomes demonstrate that the combination of Kolmogorov--Arnold representation principles and an efficient attention mechanism offers a practical and robust solution for high-resolution crop monitoring and supports scalable decision-making in precision agriculture.
\end{abstract}

\begin{keyword}
 Unmanned aerial vehicles\sep Agricultural phenotyping\sep Object localization\sep Kolmogorov-Arnold networks
\end{keyword}

\end{frontmatter}


\section{Introduction}
Maize cultivation spans approximately 200 million hectares worldwide and plays a central role in agricultural production, water resource management, and ecological sustainability \citep{background1, maizeloc2}. High-resolution unmanned aerial vehicle (UAV) imagery has become an important tool for precision agriculture, offering centimeter-level spatial detail that enables point-level plant monitoring and scalable field assessment. The increasing availability of operational UAV systems has supported widespread adoption of fine-resolution crop sensing for data-driven agronomic decision-making \citep{background5, UAV2}.

Despite these advances, point-level maize localization in UAV imagery remains technically challenging. Maize plants often occupy less than 0.1\% of an image, creating extremely low target-to-background ratios that complicate small-object detection. At the same time, processing high-resolution imagery with millions of pixels imposes substantial memory and runtime demands, especially for real-time or near-real-time agricultural applications. Additional complexities—including sparse plant distributions, heterogeneous field conditions, and visual variability across growth stages—further reduce the robustness of traditional feature extraction and localization strategies \citep{cnnloc1, flyplan}.

To address these challenges, crop monitoring techniques have evolved from manual field surveys to satellite-based analyses and, more recently, to UAV-based high-resolution imaging \citep{traditionalloc, UAV1}. While conventional computer vision approaches relying on handcrafted features have provided useful insights, deep learning has substantially advanced agricultural localization performance. Convolutional neural network (CNN)–based methods, in particular, have demonstrated strong capabilities in crop detection, counting, and spatial estimation tasks under diverse field environments \citep{cnnloc, SCALNet}.

Building upon these foundations, Visual Transformers have emerged as a particularly promising approach through innovative architectures such as GeoFormer's separable perceptron modules and FSRA's aggregation window attention mechanisms, successfully capturing long-range spatial dependencies crucial for understanding complex crop row structures and field-level patterns \citep{GeoFormer, UAVformer}. These advances enable automated extraction of spatially-explicit agricultural information essential for precision farming and yield monitoring applications.

However, despite these substantial advances in agricultural computer vision, direct application of standard Transformers to agricultural contexts faces significant computational bottlenecks. Intensive Multi-Layer Perceptrons (MLP) and quadratic complexity self-attention mechanisms limit practical deployment on resource-constrained agricultural platforms, particularly for real-time photogrammetric processing workflows \citep{VIT, Separableatt}.

In response to these computational limitations, recent innovations have explored alternative neural architectures specifically suited for agricultural applications. Notably, Kolmogorov-Arnold Networks (KAN) have emerged as compelling alternatives to traditional MLPs, decomposing complex multivariate functions into interpretable univariate components that align well with the hierarchical nature of agricultural scenes—from individual plants to field-level patterns \citep{KAN, KANeff}. Concurrently, efficient attention mechanisms have been developed to address self-attention's quadratic time complexity while preserving the ability to model global spatial relationships essential for crop localization \citep{Reformer, Separableatt}. Nevertheless, agricultural point localization presents unique challenges where high-resolution imagery generates excessive token counts while sparse object distributions characteristic of crop fields result in computational inefficiency when processing vast background regions \citep{cnnloc1, CLTR}.

\begin{figure}[t]
	\centering
	\includegraphics[width=1\columnwidth]{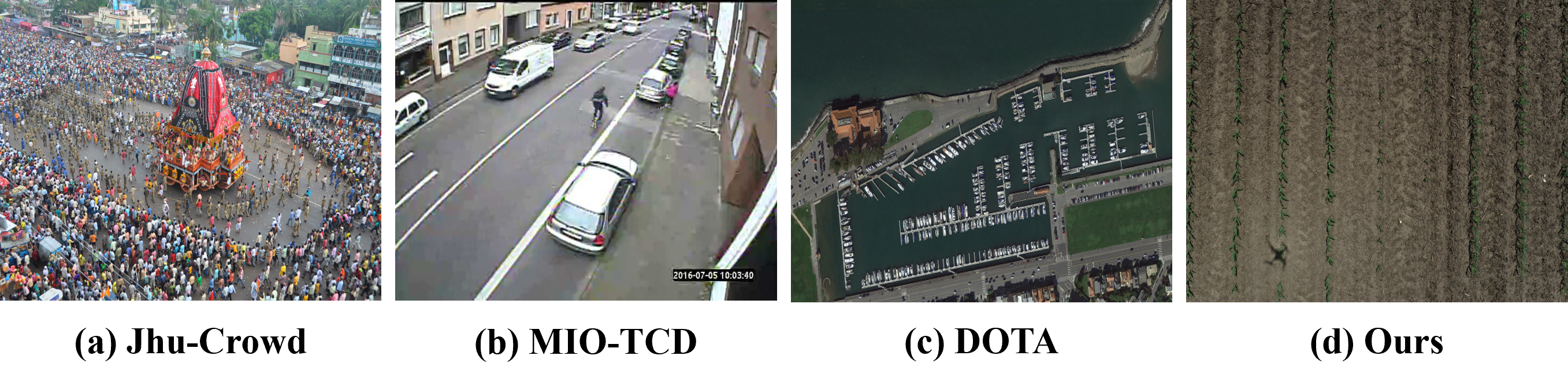}
    \vspace{-0.8cm}
	\caption{Various localization datasets are shown from left to right: JHU-Crowd\citep{Jhu-crowd}, MIO-TCD\citep{MIO-TCD}, DOTA\citep{DOTA}, and the proposed PML dataset (detailed in Section\ref{sec:dataset}). Unlike conventional localization datasets, PML is specifically designed for real-world maize scenarios, providing precise annotations tailored to agricultural scenarios. As observed in the images, real-world data presents several challenges, including a small object-to-pixel ratio, highly variable field conditions, and significant background noise, all of which make accurate model localization more challenging and demand robust feature extraction and learning strategies.}
	\label{fig:datasets}
\end{figure}

Beyond these algorithmic challenges, agricultural remote sensing faces an equally critical data limitation: the scarcity of point-level annotated datasets. While comprehensive datasets exist for general object detection using bounding box annotations \citep{Jhu-crowd, DOTA}, agricultural-specific resources with precise point annotations remain severely limited, as illustrated in Table~\ref{tab:datasets} and Figure~\ref{fig:datasets}.

\begin{table}[t]
\centering
\caption{Overview of publicly available localization datasets. The symbol "$\sim$" denotes approximate values. Point-Level annotations indicate instances labeled with a single point, whereas Box-Level annotations use bounding boxes to define object boundaries. \emph{Avg.} resolution represents the average image resolution within each dataset.}
\vspace{0.1cm}
\resizebox{0.9\linewidth}{!}{ 
\begin{tabular}{c|c|c|c|c}
\toprule
\textbf{Dataset} & \textbf{Type} & \textbf{Resolution} & \textbf{Annotations} & \textbf{\# of Images/Annotations} \\
\midrule
JHU-CROWD\citep{Jhu-crowd} & Various & $910 \times 1430$& Point-/Box-Level & $4,372\sim  \,/\, \sim 1,510,000$ \\
MIO-TCD\citep{MIO-TCD} & Camera & $720 \times 1080$; $1080 \times 1920$ & Box-Level & $648,959\,/\, -$ \\
NWPU\citep{nwpu} & Camera & $720 \times 1080$; $1080 \times 1920$ & Point-/Box-Level & $94,986\,/\,2,133,375 $ \\
AgrV\citep{AgrV} & Aerial & $512 \times 512$ & Pixel-Level & $5,109\,/\, 37,000,000$ \\
DOTA \citep{DOTA} & Aerial & Avg. $4000 \times 4000$ & Box-Level & $11,268 \,/\, 1,793,658$\\
DeepFish \citep{deepfish} & Camera & $1080 \times 1920$ & Point-Level & $40,000 \,/\, \sim 22,400$ \\
TasselNetv3 \citep{tasselnetv3} & UAV & $3648 \times 5472$ & Point-Level & $400 \,/\, 70,870$ \\
PML (Ours) & UAV & $3648 \times 4864$ & Point-Level & $1,928 \,/\, \sim 501,280$ \\
\bottomrule
\end{tabular}}
	\label{tab:datasets}
\end{table}

As revealed in Table~\ref{tab:datasets}, among existing localization datasets, only TasselNetv3 provides UAV-based point annotations for agricultural applications, but with limited scale. Furthermore, Figure~\ref{fig:datasets} demonstrates the unique challenges posed by real-world agricultural scenes, including small object-to-pixel ratios, variable field conditions, and significant background noise that distinguish agricultural datasets from conventional localization benchmarks.

This annotation gap is particularly problematic for crop localization tasks where point-level geometric precision is essential for spatial analysis applications such as plant counting, spacing estimation, and yield mapping. Current agricultural datasets predominantly employ computationally expensive bounding box or segmentation annotations that require extensive manual effort while constraining the development of lightweight point-based models essential for real-time field deployment \citep{tasselnetv3}.

To overcome these multifaceted challenges simultaneously, we present the Additive Kolmogorov-Arnold Transformer (AKT), which introduces two key innovations for efficient agricultural object localization with enhanced spatial analysis capabilities. First, we propose Padé KAN (PKAN) modules that replace conventional MLPs with learnable rational functions, offering superior functional approximation capabilities specifically suited for modeling complex agricultural scene patterns while maintaining computational efficiency. Second, we develop PKAN Additive Attention (PAA) mechanisms that substitute quadratic self-attention with efficient additive operations, enabling effective processing of high-resolution imagery without computational explosion.

To address the data limitation, point-level annotations offer distinct advantages over bounding box approaches for geospatial applications: they require significantly less annotation effort, enable more accurate density estimation for counting tasks, support precise spatial measurements for agronomic analysis, and facilitate lightweight model architectures essential for real-time UAV deployment. Therefore, we additionally introduce the Point-based Maize Localization (PML) dataset, comprising 1,928 UAV images with approximately 501,000 point annotations collected under diverse field conditions with rigorous geometric accuracy validation.

Comprehensive experimental validation demonstrates that AKT achieves an average F1-score of 62.8 percent, representing a 4.2 percent improvement over state-of-the-art baselines, while reducing computational requirements by 12.6 percent FLOPs and increasing inference throughput by 20.7 percent. Moreover, the framework demonstrates sub-2 cm geometric accuracy for plant spacing estimation, meeting the precision requirements for operational precision agriculture applications.

In summary, the main contributions of this work are:
\begin{itemize}
\item \textbf{Architectural Innovation}: The AKT framework integrates Padé KAN modules and additive attention mechanisms, achieving 4.2 percent improvement in F1-score while reducing computational complexity by 12.6 percent in Floating Point Operations compared to baseline Transformers.
\item \textbf{Dataset Development}: The PML dataset provides the largest publicly available collection of point-annotated agricultural imagery, enabling standardized evaluation of localization algorithms under realistic field conditions.
\item \textbf{Practical Applications}: Beyond localization, AKT demonstrates versatility in downstream tasks including stand counting and interplant spacing estimation, achieving 20.7 percent higher throughput for real-time deployment.
\end{itemize}

The remainder of this paper is organized as follows. Section~\ref{sec:method} first presents the PML dataset construction and AKT architecture. Then, Section~\ref{sec:experiments} details experimental protocols, ablation studies, and comparative analyses. Section~\ref{sec:discussion} discusses limitations and potential improvements. Finally, Section~\ref{sec:conclusion} concludes with key findings and future research directions.

\section{Proposed Method}
This section presents the proposed Additive Kolmogorov-Arnold Transformer framework for point-level maize localization. Section~\ref{sec:method} introduces the overall architecture and core components, including the PKAN modules and PAA mechanisms. Section~\ref{loss} formulates the training objective. Section~\ref{sec:dataset} describes the dataset construction process.

\subsection{Dataset Construction}
\label{sec:dataset}

\textbf{Data Acquisition.} To address the scarcity of annotated datasets tailored for maize localization in agricultural remote sensing, a new dataset was curated specifically for point-based localization tasks. The dataset comprises 483 high-quality UAV images, carefully selected from an initial pool of 1,142 raw images through rigorous radiometric and geometric quality assessment, and includes over 501,280 manually annotated maize locations, as summarized in Table~\ref{tab:data}.

\begin{table}[ht]
    \centering
    \caption{Overview of the point-based maize core localization dataset, including the number of original and filtered images along with their respective resolutions.}
    \vspace{0.1cm}
    \label{tab:data}
    \resizebox{0.7\columnwidth}{!}{
    \begin{tabular}{lccc}
    \hline
    Date & \# Original Images & \# Filtered Images & Resolution \\ 
    \hline
    June 12 & 372 & 185 & 3648 $\times$ 4864 \\
    June 14 & 354 & 177 & 3648 $\times$ 4864 \\ 
    June 15 & 177 & 85 & 3648 $\times$ 4864 \\
    June 17 & 239 & 36 & 3648 $\times$ 4864 \\ 
    \hline
    Total & 1,142 & 483 & 3648 $\times$ 4864 \\ 
    \hline
    \end{tabular}
    }
\end{table}

All images were captured using a DJI Phantom 4 Pro V2.0 UAV equipped with a high-resolution RGB camera (20 MP, 1-inch CMOS sensor) and a three-axis mechanical gimbal for image stabilization. The UAV flights were conducted at 120 m above ground level, achieving a Ground Sampling Distance (GSD) of 3.1 cm with 80\% forward overlap and 70\% side overlap to ensure complete stereoscopic coverage and eliminate data gaps. Flights were executed at the Arlington Agricultural Research Station ($43^{\circ}18'\mathrm{N},\, 89^{\circ}24'\mathrm{W}$), University of Wisconsin--Madison, USA, following pre-programmed flight paths with waypoint navigation to maintain consistent geometric configuration across multiple acquisition dates.

\textbf{Data Quality Control.} All imagery was acquired under clear sky conditions, with cloud coverage $<$ 10\% and solar elevation angles $>30^{\circ}$ to minimize shadow effects and ensure uniform illumination. Flight missions were conducted between 10:00 AM and 3:00 PM local time to optimize solar geometry and reduce atmospheric scattering effects. Wind speeds were maintained below 8 m/s during acquisition to minimize motion blur and geometric distortions. The atmospheric visibility exceeded 15 km on all flight days (June 12, 14, 15, and 17, 2022), ensuring minimal atmospheric attenuation.

\begin{figure}[t]
\centering
\includegraphics[width=1\columnwidth]{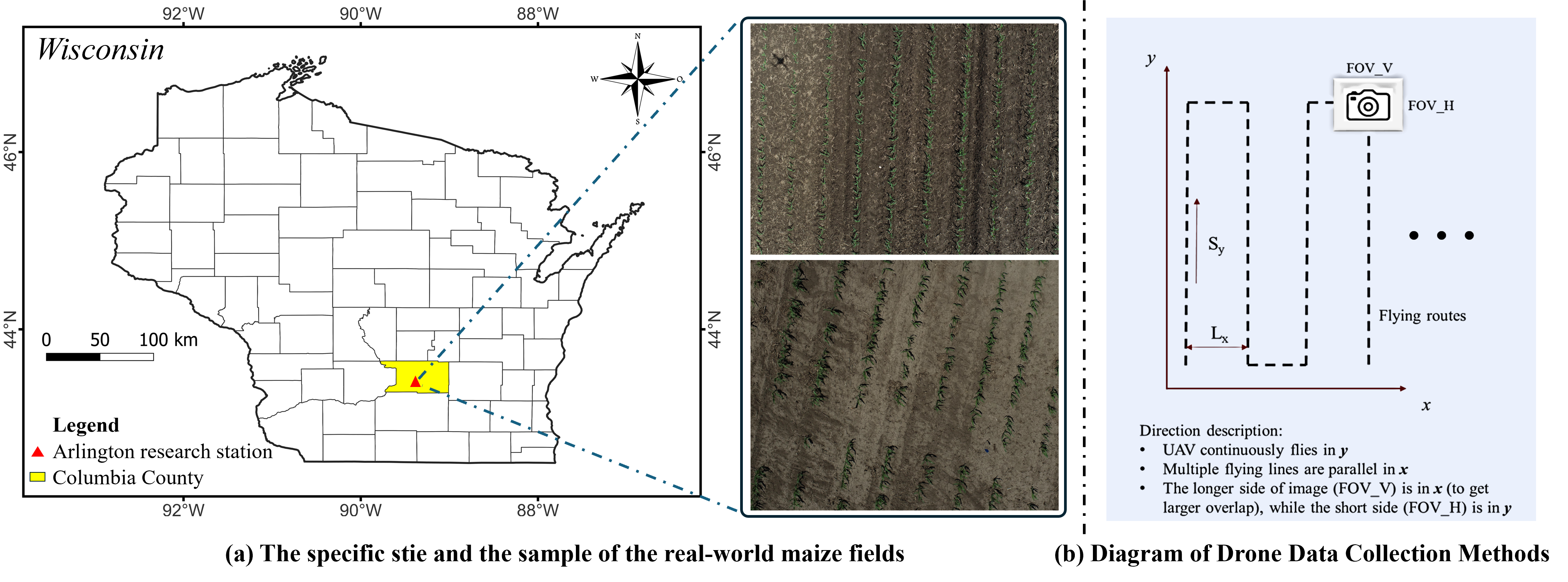}
\caption{The specific location for the UAV imagery collection of real maize data, along with representative sample images and definitions of UAV flight paths and associated parameters, are detailed in Section~\ref{sec:dataset}. (a)The specific site and the sample of the real-world maize fields. The red triangular marker indicates the actual collection site at the Arlington Research Station. (b) Diagram of Drone Data Collection Methods. The Unmanned Aerial Vehicle (UAV) follows a preplanned flight path to cover the entire experimental area, ensuring the completeness of data collection.}
\label{fig:field}
\end{figure}

To ensure both radiometric consistency and geometric accuracy, each image underwent rigorous quality screening using remote sensing standards. An example of aerial coverage and flight parameters is illustrated in Fig.~\ref{fig:field}(b). Note that the field experiments were conducted as part of a genome-to-phenome (GxE) initiative evaluating hybrid maize performance under varying spatial configurations. Therefore, in the experimental field where the imagery was collected, each plot adopted a two-row layout, with rows measuring 6.7 meters in length and spaced 0.97 meters apart. Approximately 80 seeds were planted per plot at 0.15-meter intervals.

\textbf{Remote Sensing Data Preprocessing.} Following standard remote sensing preprocessing protocols, all images underwent geometric and radiometric corrections before analysis. Geometric preprocessing included lens distortion correction using camera calibration parameters and orthorectification to compensate for terrain relief and viewing angle variations. Radiometric preprocessing involved digital number to radiance conversion using sensor-specific calibration coefficients, followed by atmospheric correction to retrieve surface reflectance values.

\begin{figure}[t]
\centering
\includegraphics[width=1\columnwidth]{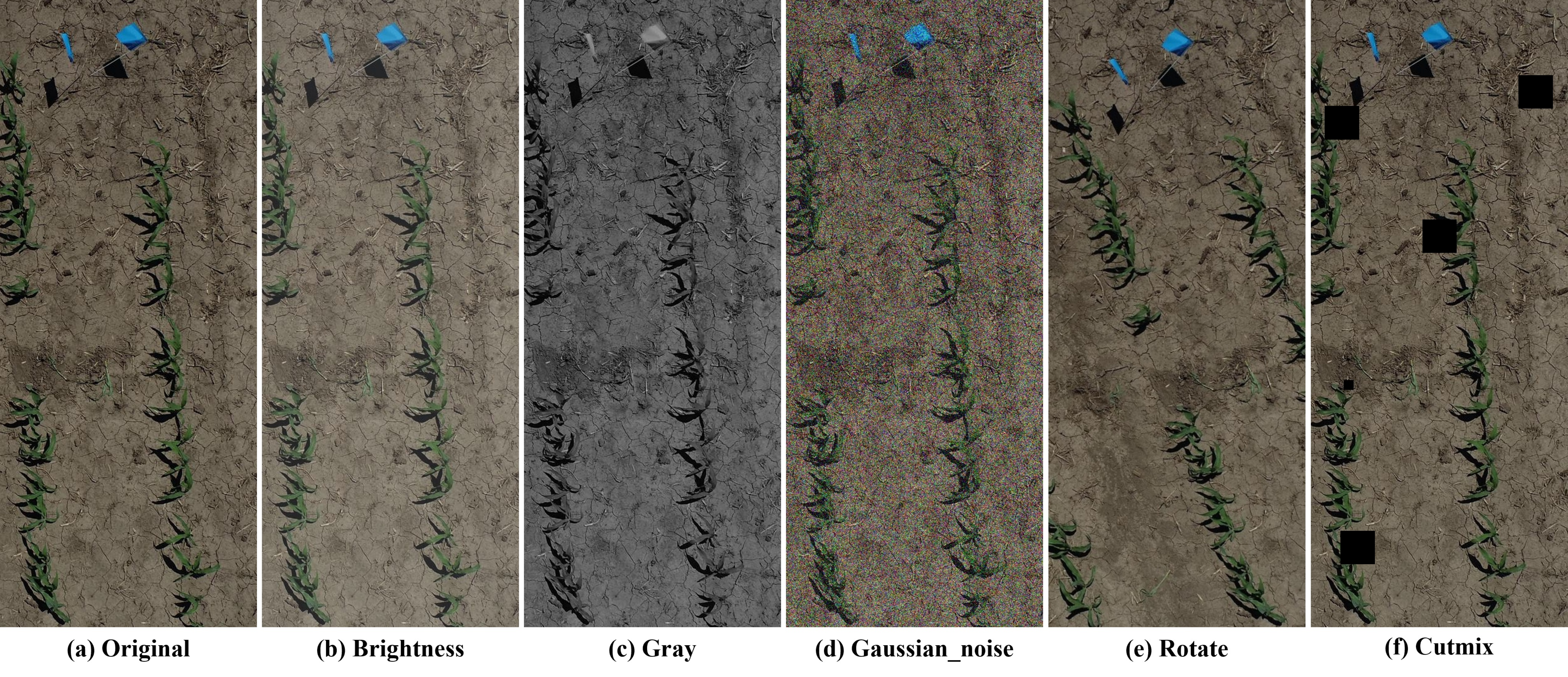}
\caption{The sample of the point-based maize localization dataset includes examples enhanced with various augmentation techniques. These techniques include brightness adjustment, grayscale transformation, the addition of Gaussian noise, rotation, and cutmix, among others. These augmentations aim to increase the diversity of the dataset, helping to improve model generalization and performance under varying real-world conditions.}
\label{fig:augmentation}
\end{figure}

To enhance model generalizability and account for natural variations in agricultural remote sensing data, physics-informed data augmentation was employed. Applied transformations included: (1) geometric augmentations simulating viewing angle variations, such as, random rotations between -15° and 15°, horizontal and vertical flipping, (2) radiometric augmentations mimicking atmospheric and illumination conditions, such as, contrast adjustment by 0.3, brightness variations, (3) sensor noise simulation through additive Gaussian noise, and (4) spatial resolution degradation and enhancement to simulate different flight altitudes. In this study, the CutMix technique was adapted to better preserve spatial context~\cite{cutmix}. Examples of these remote sensing-specific augmentations are shown in Fig.~\ref{fig:augmentation}.

Following the distribution strategy of the NWPU-crowd dataset, the PML was divided into training, validation, and testing subsets in a 60/15/25 ratio. The final distribution of the UAV image of maize after pre-processing and augmentation is summarized in Table~\ref{tab:dataset}. In addition, agricultural experts manually verified the annotations to ensure the reliability and precision of the annotation data.

\begin{table}[t]
    \centering
    \caption{Division results of PML dataset.}
    \vspace{0.1cm}
    \label{tab:dataset}
    \resizebox{0.7\columnwidth}{!}{
    \begin{tabular}{ccccc}
    \hline
    &\# Images &\# Training  &\# Validation  &\# Testing  \\ \hline
    Filtered   & 483   & 289   & 72   & 122   \\ 
    Augmented  & 1,928  & 1,734  & 72   & 122   \\ 
    \hline
    \end{tabular}
    }
\end{table}

\subsection{Network Architecture}
\label{sec:method}

\subsubsection{Architecture Overview}
To meet the requirements of high-precision and computationally efficient maize localization in agricultural remote sensing applications, the AKT framework is proposed. The method specifically addresses two challenges inherent in agricultural remote sensing: i) the need for high spatial resolution to detect individual plants in centimeter-level GSD imagery, and ii) computational constraints associated with real-time deployment in precision agriculture localization systems.

\begin{figure*}[t]
\centering
\includegraphics[width=1\columnwidth]{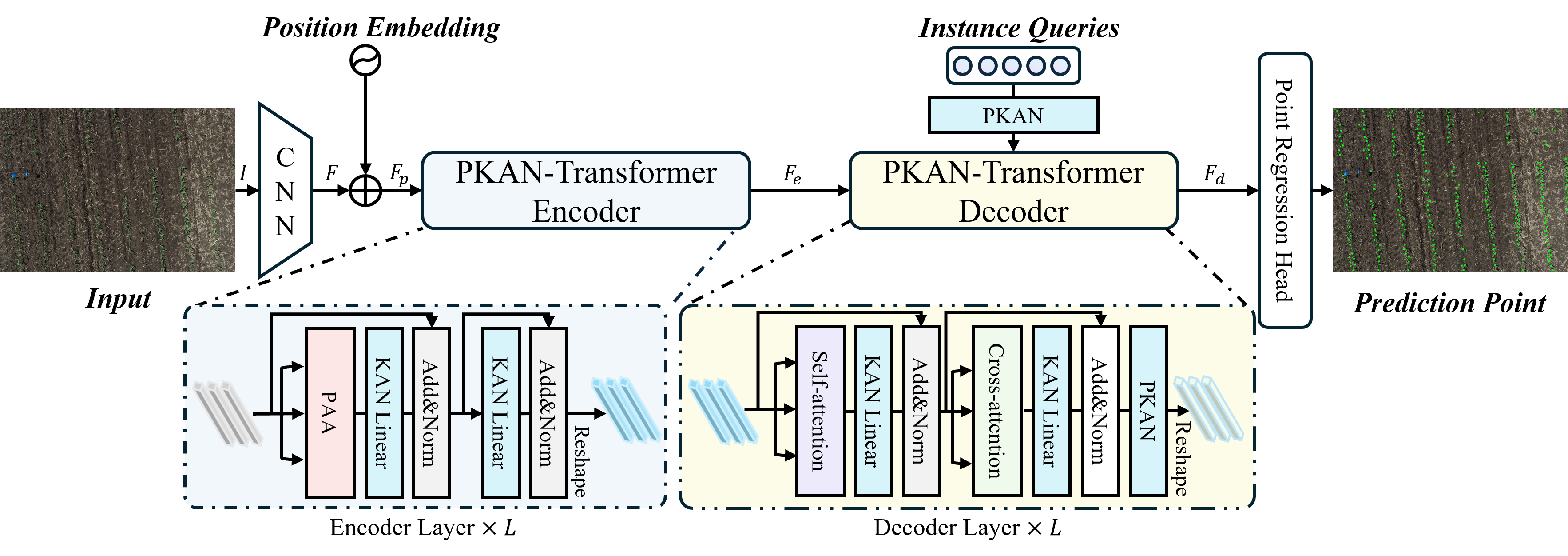}
\caption{Overview of the proposed Additive Kolmogorov-Arnold Transformer (AKT) network. The AKT consists of multiple encoder and decoder layers.
First, a CNN-based backbone is used to extract feature maps $F$ from the input image. Then, $F$ are added with positional embeddings, generating $F_{p}$, which is subsequently fed into the PKAN-Transformer encoder. The encoder captures contextual information, aggregates spatial details, and outputs the encoded features $F_{e}$. The encoded output features $F_{e}$, along with instance queries processed by the PAA, KAN layer, and other components, are passed through various attention modules and PKAN modules in the decoder. This process produces the decoder output $F_{d}$, which is used to predict the maize core.}
\label{fig:overall}
\end{figure*}

The AKT adopts a modified encoder-decoder paradigm optimized for remote sensing data characteristics, comprising two main components: a CNN-based backbone (e.g., \emph{ResNet50}) adapted for high-resolution agricultural imagery feature extraction and a modified Padé KAN (PKAN) Transformer encoder-decoder network designed to handle the spatial sparsity and scale variations typical in crop monitoring applications. The overall architecture of AKT is illustrated in Fig.~\ref{fig:overall}.

\textbf{Encoder Component.} The encoder comprises several layers specifically designed for agricultural remote sensing data, each containing a PKAN Additive Attention (PAA) module and two KAN linear layers. Its primary function is to capture multi-scale contextual information critical for crop row structure understanding and aggregate spatial details across varying GSD conditions. The encoder leverages KAN's mathematical function approximation capabilities to model complex spectral-spatial relationships in agricultural scenes, processing global field context while maintaining sensitivity to individual plant signatures that may vary across different illumination and atmospheric conditions.

\textbf{Decoder Component.} The decoder consists of multiple stacked layers optimized for detecting sparse targets in agricultural scenes, each of which includes two PKAN modules, a self-attention module, a cross-attention module ~\cite{cdetr} and two linear layers of KAN. Its role is to manage complex spatial dependencies typical in crop fields (such as regular planting patterns, row structures, and plant spacing variations) and predict precise sub-pixel localization coordinates. The decoder architecture accounts for the inherent spatial autocorrelation in agricultural fields while maintaining robustness to geometric distortions and radiometric variations common in UAV-acquired imagery.

\textbf{Pipeline.} Given an input image $I \in \mathbb{R}^{H\times W\times 3}$, where $H$ and $W$ denote the height and width, a convolution-based backbone first extracts feature maps $F \in \mathbb{R}^{\frac{H}{32} \times \frac{W}{32}\times C}$. These features are flattened into a sequence with positional embeddings, reducing the channel dimension from $C$ to $c$, generating $F_{p} \in \mathbb{R}^{\frac{H}{32} \times \frac{W}{32}\times c}$.

The encoder takes \(F_{p}\) as input and produces the encoded features \(F_{e}\), which capture the global context of the scene. These features are then processed through multiple decoder layers, where PKAN modules generate a refined set of candidate representations.  To enhance computational efficiency and reduce memory overhead, the traditional reliance on a large number of predefined candidates is replaced with learnable instance queries.These queries focus directly on potential targets, enabling more effective feature interactions and optimizing resource utilization while maintaining localization accuracy.

During decoding, the learnable instance queries interact with the encoded features to generate the final decoded embeddings \(F_{d}\). This structured approach ensures that the model efficiently captures critical spatial information without incurring excessive computational costs. Finally, the decoded features \(F_{d}\) are passed through a point regression head, which predicts the maize core coordinates with high precision.

\subsubsection{Padé KAN (PKAN) Module}
Agricultural remote sensing data presents unique challenges that standard MLPs struggle to address effectively. These challenges include: (1) complex spatial interactions between crop canopy structure and sensor viewing geometry, and (2) multi-scale feature representations required for detecting objects ranging from individual plants to field-level patterns ~\cite{pkan1, pkan2}. Directly replacing MLPs with standard KAN layers, while theoretically promising, proves insufficient for handling these remote sensing-specific complexities efficiently.

To address these limitations and optimize performance for agricultural remote sensing applications, the Kolmogorov-Arnold representation is revisited and the standard KAN framework is analyzed to assess its expressiveness and suitability for integration with Transformer-based architectures. In its conventional form, KAN consists of a composition of \(L\) layers~\cite{KAN}. Given an input vector $x_{0} \in \mathbb{R} ^{n_{0}}$, the output is:
\begin{equation}
\mathrm{KAN}(\mathbf{x})=(\boldsymbol{\Phi}_{L-1}\circ\boldsymbol{\Phi}_{L-2}\circ\cdots\circ\boldsymbol{\Phi}_{1}\circ\boldsymbol{\Phi}_{0})\mathbf{x}
\end{equation}
where $\boldsymbol{\Phi}_{L}$ is the function matrix corresponding to the $l$-th KAN layer, defined as a matrix of 1D functions:
\begin{equation}
\boldsymbol{\Phi} = \{\phi_{q,p}\}, \quad p = 1, 2, \ldots, n_{in}, \quad q = 1, 2, \ldots, n_{out}
\end{equation}
where the function $\phi$ has trainable parameters $n_{in}$ dimensional input and $n_{out}$ dimensional output:
\begin{equation}
\phi(x) = w_b \texttt{silu}(x) + w_s \texttt{spline}(x)
\end{equation}
where $\texttt{spline}(x)$ is parameterized as a liner combine of $\texttt{B-splines}$ function ($B_{i}$) as $\texttt{spline}(x)=\sum_{i}c_{i}B_{i}(x)$, and $c_{i}$ are trainable which implies the model can dynamically learn and adjust the $\texttt{spline}$’s shape during the training process~\cite{KAN}.

While $\texttt{SiLU}$ has demonstrated promising performance across various tasks, its relatively high computational cost and potential gradient instability under certain conditions make it less suitable for lightweight or real-time applications \cite{silu, KAN, silu1}. Given the critical role of expressiveness and precision in capturing boundary details and identifying small objects, the $\texttt{SiLU}$ activation function becomes a bottleneck in localization tasks due to its inefficiency. To address this limitation and inspired by~\cite{PAU}, the Padé KAN (PKAN) module replaces $\texttt{SiLU}$ with the safe and efficient Padé Activation Unit (PAU), thereby enhancing both the accuracy and computational efficiency of the original KAN framework~\cite{KAN}.

The modified PKAN leverages compact representations and improves computational efficiency by utilizing flexible parametric rational functions that approximate commonly used activation functions. This design reduces redundant parameters and ensures greater compatibility.
Moreover, the Padé activation unit enables the optimization process to automatically determine the appropriate activation function for each layer during training, enhancing adaptability and expressiveness. 
In the proposed PKAN module, the PAU replaces the fixed activation function \(f(x)\), while the rational function \(F(x)\) and \(\phi(x)\) on each edge are fully parameterized, offering greater flexibility and efficiency in the real-world maize core localization task.
Formally,
\begin{equation}
\phi(x)=wF(x)=w\frac{P(x)}{Q(x)}=w\frac{a_0+a_1x+\cdots+a_mx^m}{1+|b_1x+\cdots+b_nx^n|}
\end{equation}
where \(P(x)\) and \(Q(x)\) are the polynomials of \(F(x)\) of the order \(m\), \(n\). PKAN can achieve fine-grained function approximation and interpolation for \(f(x)\) when integrated into the Transformer architecture.

In this study, the proposed PKAN module is formulated as a specialized MLPs characterized by three key components: (1) learnable nonlinear transformations, (2) the incorporation of a rational function prior to the linear layer, and (3) group-specific activation functions tailored to different edge categories. As shown in the ablation experiments and summarized in Table~\ref{tab:kan}, PKAN achieves superior performance over other KAN variants in the maize core localization task in Section~\ref{abl_PKAN}. The performance gain is primarily attributed to the integration of the PAU with rational functions, in which the denominator coefficients are shared among the groups, while the numerator coefficients are learned independently.  This design achieves an effective balance between model complexity and expressive capacity, making PKAN particularly suitable for deployment on resource-limited platforms, such as mobile devices and UAVs, for agricultural remote sensing applications.

\subsubsection{PKAN Additive Attention Mechanism}
This section presents the PAA, which is developed to address the issue of sample sparsity in high-resolution UAV-based localization tasks. In contrast to previous approaches that primarily refine self-attention mechanisms, such as, through deformable attention~\cite{detr} or hybrid designs with convolutional layers~\cite{drcnet, tbaie} to reduce computational burden, the proposed method adopts an alternative strategy. Specifically, PAA employs additive operations combined with PKAN, thereby enhancing computational efficiency while maintaining representational capability.

\begin{figure}[t]
\centering
\includegraphics[width=1\columnwidth]{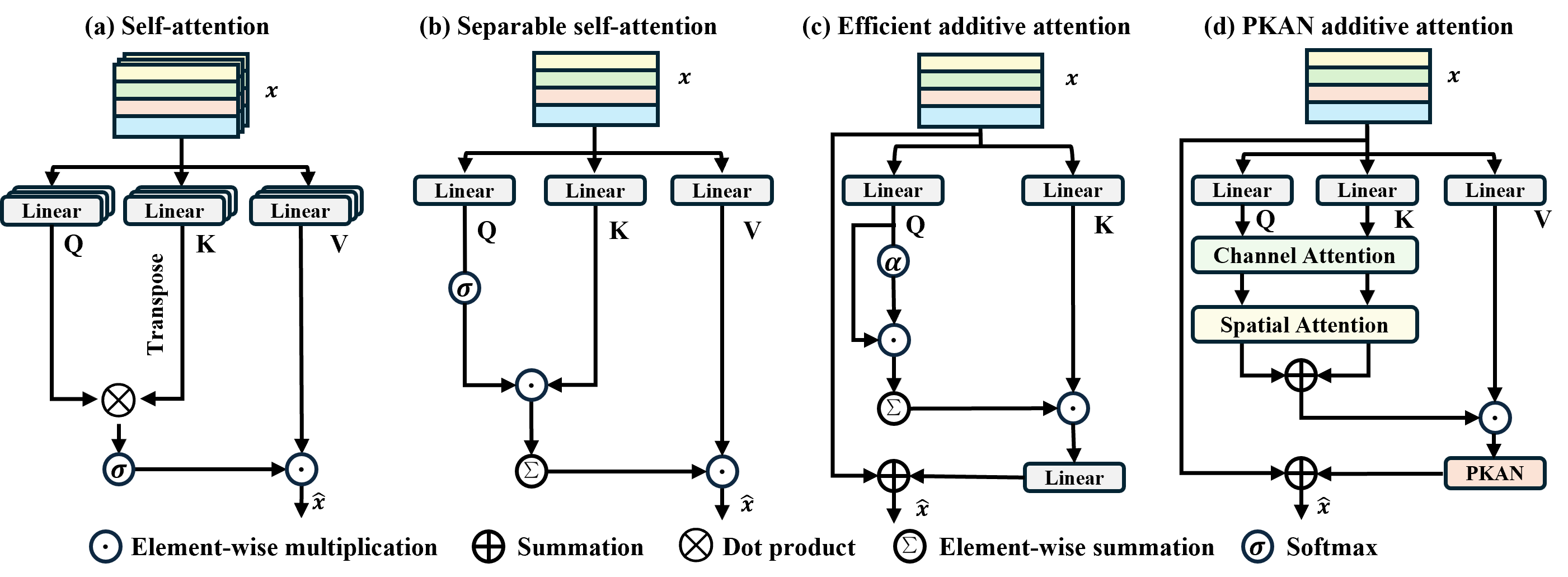}
\caption{Comparison of various self-attention module variants: (a) represents the classical self-attention mechanism in ViT~\cite{VIT}; (b) shows the separable self-attention from MobileViTv2~\cite{Separableatt}, which simplifies the feature representation by reducing a matrix to a vector; (c) illustrates the swift self-attention from SwiftFormer~\cite{Swiftformer}, employing additive attention to eliminate costly matrix multiplication operations; and (d) presents the proposed PKAN Additive Attention (PAA) module.
Unlike previous attention mechanisms, PAA replaces the MLP with PKAN while maintaining accuracy. Furthermore, the key-value interaction leverages additive operations, effectively avoiding the quadratic computational complexity typical of traditional self-attention mechanisms.}
\label{fig:PAA}
\end{figure}

Existing attention mechanisms are based on dense interactions, which inherently constrain the spatial scope of content queries~\cite{cdetr, drcnet} and limit the extraction of global features. Such restrictions reduce the representational capacity and adversely affect the performance of maize localization in UAV imagery. To address this issue, the proposed PAA module leverages the expressiveness of PKAN to emphasize key relationships and operates effectively under data sparsity. Drawing inspiration from separable self-attention~\cite{Separableatt} and Swiftformer~\cite{Swiftformer}, PAA reformulates the Multi-Head Attention (MHA) mechanism by integrating PKAN with additive operations to mitigate quadratic complexity. This design avoids computationally intensive components, such as matrix multiplications and MLPs, thereby improving efficiency. The structure of PAA is illustrated in Fig.~\ref{fig:PAA}.

Similar to conventional MHA, the proposed PAA takes an input $x \in \mathbb{R}^{N \times d}$, which is processed through three linear projections to obtain the Query ($Q$), Key ($K$), and Value ($V$) representations. Here, $N$ denotes the number of tokens and $d$ represents the embedding dimension within each head. To enhance global information and strengthen feature representation~\cite{Cbam}, Channel Attention $CA(\cdot) \in \mathbb{R}^{N \times d}$ and Spatial Attention $SA(\cdot) \in \mathbb{R}^{N \times d}$ are applied to $Q$ and $K$, respectively.

Subsequently, PAA introduces a summation-based context mapping function to compute the context vector, enabling a more computationally efficient estimation of global dependencies. This mechanism is conceptually analogous to the similarity function employed in self-attention. The resulting context mapping is then multiplied by the value matrix $V$, effectively capturing position-wise relationships. The final output of PAA, denoted as  $y \in \mathbb{R} ^{N\times d}$, is computed as follows:
\begin{equation}
y = Concat\left ( SA(CA(Q)),SA(CA(K)) \right ) \cdot V
\end{equation}

The proposed PKAN is used to improve the model's representational capacity and improve its nonlinear modeling capabilities. The resulting transformation is given by:
\begin{equation}
\hat{x} = x + PKAN(y)
\end{equation}

The PAA mechanism operates through three key stages optimized for remote sensing applications: i) \textit{Spatial decomposition}: Input features are decomposed into local plant-level components and global field-level components based on the characteristic spatial scales in agricultural scenes; ii) \textit{Additive attention computation}: Attention weights are computed additively rather than multiplicatively, reducing complexity from $O(N^2)$ to $O(N)$ while preserving the ability to model long-range dependencies across crop rows; iii) \textit{PKAN-enhanced feature interaction}: PKAN modules are integrated within attention computation to model complex non-linear relationships between various spatial patterns in agricultural remote sensing data.

Additionally, the PAA mechanism provides an optimized representation for remote sensing tasks by substantially reducing computational complexity while preserving the capacity to capture both fine-scale plant morphology and large-scale field structures. This makes it particularly suitable for operational deployment in precision agriculture systems relying on high-resolution UAV imagery. Moreover, the mechanism explicitly accounts for the inherent anisotropy of agricultural scenes, characterized by stronger correlations along crop rows than across rows, and adapts the attention computation accordingly.

\subsection{Loss Function}
\label{loss}
To train the proposed AKT model, predictions are matched to Ground Truth (GT) points in a one-to-one manner. This is formulated as a bipartite matching problem between predicted points $P_{i}$ and GT points $G_{j}$, where the $\mathcal{L}{1}$ distance is employed to measure the matching cost:
\begin{equation}
 \mathcal{L} =\left \| G_{j}- P_{i} \right \| _{1} -\hat{C}_{j} , \quad i \in M, j \in N
\label{eq:l1}
\end{equation}
where $P_{i}$ denotes the coordinates of the $i$-th prediction and $G_{j}$ the coordinates of the $j$-th GT point. $M$ and $N$ represent the sets of predicted and GT points, respectively, with $|N| > |M|$ to ensure that each GT point is assigned to a prediction, while unmatched predictions are treated as negatives. The matching process minimizes the total cost using an $\mathcal{L}{1}$-based Hungarian algorithm~\cite{cdetr}, with $\hat{C}{j}$ denoting the confidence score associated with the $j$-th predicted point.

\section{Experimental Results}
\label{sec:experiments}
This section provides a comprehensive evaluation of the proposed AKT framework in the context of agricultural remote sensing applications. Section~\ref{Experiment} describe the experimental setup, including implementation details, baseline methods, and evaluation metrics. Section~\ref{performance} then presents a comparative analysis against state-of-the-art approaches, highlighting the superior performance and enhanced capabilities of AKT for agricultural remote sensing tasks. Finally, Section~\ref{ablation} reports systematic ablation studies, quantifying the contribution of each AKT component to the overall performance and validating the effectiveness of the proposed design.

\subsection{Experimental Settings}
\label{Experiment}

\textbf{Implementation Details.} 
The experiments were conducted following established remote sensing validation protocols on the high-resolution agricultural PML dataset. The computational infrastructure consisted of an NVIDIA H100 Tensor Core GPU with 80 GB memory and an Intel Xeon Platinum 8460Y+ processor, optimized for large-scale geospatial data processing. The software environment was based on Python and PyTorch~\cite{pytorch}, configured to ensure high-precision geometric processing and radiometric consistency across multi-temporal acquisitions.

The Adam optimizer~\cite{adam} was employed with parameters $\beta_{1}=0.9$ and $\beta_{2}=0.999$, an initial learning rate of $1 \times 10^{-5}$, and a weight decay of $1 \times 10^{-5}$~\cite{ddetr}. Training was carried out with a batch size of 32 for 1,500 epochs, with early stopping applied based on validation performance to mitigate overfitting to agricultural scene patterns.

The PML dataset comprises ultra-high-resolution UAV imagery with a GSD of 1 mm/px, derived from orthomosaic generation using the Quantum Geographic Information System (QGIS). Each raw image has a spatial resolution of $3648 \times 4864$ pixels. To reduce computational cost, all images were resized so that the longer side was limited to 2048 pixels, resulting in a final resolution of $1538 \times 2048$ pixels while preserving the original aspect ratio. This corresponds to a downscaling factor of approximately 2.38, resulting in an effective GSD of about 2.38 mm/px. The resolution remains sufficient for individual plant detection while maintaining the spatial context required for point-level localization in precision agriculture applications.

\textbf{Compared Methods.}
To rigorously assess the effectiveness of AKT, comparative experiments were conducted against a diverse set of representative baselines, including LCFCN~\cite{LCFCN}, LSC-CNN~\cite{LSCCNN}, GPR~\cite{GPR}, SDNet~\cite{SDNet}, RAZ\_Loc~\cite{liu2019recurrent}, AutoScale\_Loc~\cite{Autoscale}, GL~\cite{GL}, SCALNet~\cite{SCALNet}, TopoCount~\cite{topo}, CLTR~\cite{CLTR}, and FIDTM~\cite{FIDTM}. For all competing methods, the experimental settings and evaluation protocols strictly followed those reported in the original publications, thereby ensuring a consistent and fair comparison. This design provides a reliable benchmark for evaluating the advantages of AKT relative to state-of-the-art approaches.

\textbf{Evaluation Metrics.}  
The proposed framework is primarily evaluated on the task of maize core localization, while auxiliary assessments of plant counting (Sec.\ref{countingper}) and plant spacing analysis (Sec.\ref{placingper}) are conducted to provide a comprehensive evaluation of network performance. For quantitative evaluation on the PML dataset, three metrics are employed: Average Precision (Av.P, \%), Average Recall (Av.R, \%), and Average F1-measure (Av.F, \%)~\cite{CLTR}.  

In this article, a predicted point \(P_i\) is regarded as correctly localized if its Euclidean distance to a GT point \(G_j\) falls below a predefined threshold, in which case it is categorized as a True Positive (TP)~\cite{LCFCN}, performance is assessed across a series of distance thresholds \(\alpha \in [1,10]\) pixels. This multi-threshold strategy provides a robust characterization of localization performance by capturing model sensitivity under varying spatial tolerances, thereby ensuring reliable assessment of detection accuracy in high-resolution UAV imagery.  

Considering that the PML dataset has a fixed GSD of 2.38 mm/px, all pixel-based thresholds can be directly converted into real-world physical distances, such as, \(\alpha=10\) px corresponds to 2.38 cm. This conversion avoids scale bias caused by different input resolutions and ensures that the reported performance metrics are physically interpretable within the context of precision agriculture applications.

\textbf{Geometric Accuracy and Error Analysis.} For geometric accuracy assessment following remote sensing standards, a predicted point $P_i$ is considered successfully matched to GT point $G_j$ when the Euclidean distance falls below predefined spatial thresholds, accounting for inherent positional uncertainties in UAV-based remote sensing systems. These uncertainties arise from multiple error sources typical in agricultural remote sensing: (1) \textit{Geometric errors}: lens distortion, platform instability, and viewing angle variations contributing to systematic positioning offsets~\cite{gemerror}; (2) \textit{Radiometric variations}: atmospheric scattering, BRDF effects, and temporal illumination changes affecting feature detection consistency~\cite{raderror}; (3) \textit{Ground truth uncertainties}: manual annotation precision limitations and natural plant centroid definition ambiguity; (4) \textit{Scale-dependent errors}: pixel mixing effects at field boundaries and spatial resolution limitations for small target detection~\cite{gterror}.

The multi-threshold evaluation framework employs spatial tolerance ranges ($\alpha$) from 1 to 10 pixels, corresponding to 0.23--2.38 cm in real-world distances, given the dataset GSD of 2.38 mm/px. This design follows established remote sensing accuracy assessment practices and allows evaluation at pixel-level precision ($<$5 cm), which is critical for individual plant monitoring~\cite{rs}.

\subsection{Performance Comparison}
\label{performance}
\begin{figure}[t]
	\centering
	\includegraphics[width=1\linewidth]{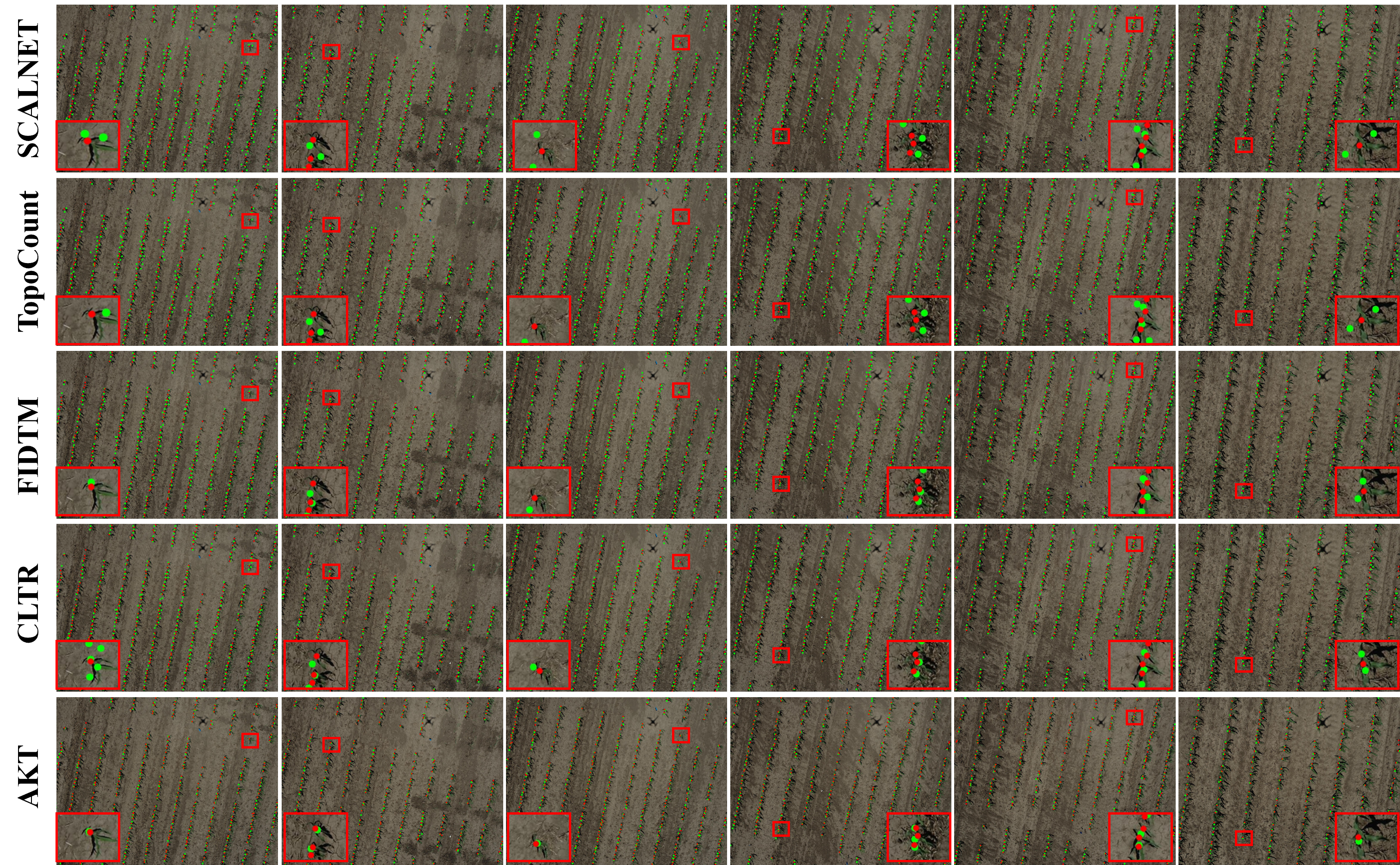}
	\caption{Qualitative visualizations of different methods on the PML dataset. Each row corresponds to a method, while columns represent different maize field samples. Green points denote predicted maize core locations, red points indicate GT annotations. To facilitate a more precise comparison, the zoomed-in regions highlight detailed localization visualizations. Compared to comparison methods, AKT demonstrates superior localization accuracy, minimizing false positives and better aligning with GT annotations, particularly in complex field conditions, showcasing its robustness for high-precision agricultural applications.}
	\label{fig:visual}
\end{figure}

The localization performance was systematically evaluated against the state-of-the-art methods on the PML dataset, with results summarized in Table~\ref{tab:localization} and Fig.~\ref{fig:visual}. This comparative analysis includes both conventional computer vision techniques and recent remote sensing-oriented approaches, providing a rigorous assessment of the proposed method’s improvements over existing agricultural monitoring strategies.

\textbf{Remote Sensing Performance Advantages.}  
In real-world agricultural remote sensing scenarios that require GSD precision of centimeters, the proposed AKT framework delivers consistent and measurable improvements. Compared with CLTR~\cite{CLTR} and FIDTM~\cite{FIDTM}, AKT achieves gains of 4.2\% on the validation set and 8.8\% on the test set in terms of average F1-score. These improvements are particularly meaningful under the operational constraints of agricultural remote sensing, where robustness to environmental variability (e.g., illumination, atmospheric turbidity, and field moisture) is critical for dependable crop monitoring and management.

\begin{table*}[t]
\centering
\caption{Localization performance evaluation on the PML dataset. Comprehensive metric analysis across spatial thresholds, reporting Average Precision (Av.P, \%), Average Recall (Av.R, \%), and Average F1-measure (Av.F, \%) at \(\alpha \in [1, 10]\) pixels~\cite{LCFCN}. The best results are highlighted in \textbf{bold}, $\uparrow$ means that a larger value is better.}
\vspace{0.1cm}
\label{tab:localization}
\scriptsize
\begin{tabular}{ccccccc}
\toprule
\multirow{2}{*}{Method} &\multicolumn{3}{c}{Validation set} &\multicolumn{3}{c}{Test set} \\ 
\cmidrule{2-7}
&Av.P (\%) $\uparrow$ &Av.R(\%) $\uparrow$ &Av.F(\%) $\uparrow$ &Av.P (\%) $\uparrow$ &Av.R(\%) $\uparrow$ &Av.F(\%) $\uparrow$ \\
\midrule
LCFCN~\cite{LCFCN}  &44.7 &41.9 &42.2 &43.9 &40.2 &41.9 \\
LSC-CNN~\cite{LSCCNN} &50.3 &46.6 &48.4 &44.3 &41.4 &42.8 \\
GPR~\cite{GPR}  &51.3 &57.2 &54.1 &52.1 &47.3 &49.6 \\
SDNet~\cite{SDNet} &55.1 &50.3 &52.6 &52.3 &45.1 &48.4 \\
RAZ\_Loc~\cite{liu2019recurrent}  &50.1 &48.3 &49.2 &55.9 &44.7 &49.7 \\
AutoScale\_loc~\cite{Autoscale} &59.3 &48.7 &53.5 &51.2 &46.8 &48.9 \\
GL~\cite{GL} &65.2 &58.4 &61.6 & 58.4 & 50.1 &53.9 \\
SCALNet~\cite{SCALNet} &64.7 &61.1 &62.9 &57.9 &52.8 & 55.2 \\
TopoCount~\cite{topo} &53.1 & 45.5 &48.4 & 50.1 &47.3  &48.7 \\
FIDTM~\cite{FIDTM} &59.8&55.9 &57.8 &57.2 &51.1 &54.0 \\
CLTR~\cite{CLTR} &68.4 &63.3 &65.8 &61.3 &56.2 &58.6 \\
AKT (Ours) &\textbf{72.3} &\textbf{67.3} &\textbf{69.8} &\textbf{65.9} &\textbf{60.3} &\textbf{62.8} \\
\bottomrule
\end{tabular}
\end{table*}

\textbf{Comparison with Traditional Remote Sensing Approaches.}  
Traditional satellite-based crop monitoring platforms, such as Landsat and Sentinel-2, provide valuable capabilities for large-area coverage and long-term temporal monitoring~\cite{oldloc,traditionalloc,traditionloc2}. However, their spatial resolutions (10–30 m GSD) fundamentally preclude the detection of individual plants. In contrast, UAV-based systems are capable of achieving centimeter-level resolution, but most existing approaches rely on conventional CNN architectures, which face significant challenges when processing ultra-high-resolution imagery ($>$3000$\times$4000 pixels), typically encountered in precision agriculture. The proposed AKT framework addresses these limitations by: 1) introducing attention mechanisms specifically designed for agricultural scene structures, 2) reducing computational complexity to support near real-time processing, and 3) ensuring robustness against geometric and radiometric variations commonly present in UAV-based remote sensing environments.

The proposed method demonstrates notable robustness under diverse agricultural remote sensing conditions, including platform instability, atmospheric variability, canopy structural heterogeneity, and the stringent sub-pixel accuracy requirements of precision agriculture decision-making. As summarized in Table~\ref{tab:runtime}, relative to CLTR, AKT achieves a 5.1\% reduction in model parameters, a 12.6\% decrease in FLOPs, and a 20.7\% improvement in throughput measured in Frames Per Second (FPS). These results underscore the efficiency and practicality of the proposed framework, highlighting its suitability for operational deployment in resource-constrained agricultural monitoring systems.

\begin{table}[h]
\centering
\setlength{\tabcolsep}{4mm}
\caption{Comparison of Computational Complexity. The experiments are conducted on an NVIDIA 4090 GPU, using input images of size $1538 \times 2048$. The best results are highlighted in \textbf{bold}.}
\vspace{0.1cm}
\label{tab:runtime}
\resizebox{0.5\columnwidth}{!}{
\begin{tabular}{cccc}
\toprule
Method  & \#Params.(M)& FLOPs(G) & FPS\\
\midrule
LSC-CNN\cite{LSCCNN} & 35.1 & 14897.58 & 3.14 \\
FIDMT\cite{FIDTM} & 66.5 & 16995.92 & 3.87 \\
TopoCount\cite{topo} & 25.8 & 3413.26 & 4.42 \\
GL\cite{GL} & 21.5 & 1727.63 & 10.22 \\
SCALNet\cite{SCALNet} & \textbf{18.6} & 511.91 & 7.09 \\
CLTR\cite{CLTR} & 43.4 & 449.12 & 19.10 \\
AKT (Ours) & 41.2 & \textbf{392.57} & \textbf{23.05} \\
\bottomrule
\end{tabular}
}
\end{table}

\subsection{Ablation Studies}
\label{ablation}
In this section, a series of ablation experiments are conducted to rigorously evaluate the contribution of each core component within the AKT framework.  

\subsubsection{Impact of Individual Components}
To systematically evaluate the effectiveness of the proposed architecture, the incremental contributions of individual modules are analyzed, with results summarized in Table~\ref{tab:components}. In particular, within the backbone architecture~\cite{cdetr}, conventional linear layers are replaced with KAN-based linear layers, standard MLP blocks are replaced with the PKAN modules, and self-attention layers are replaced with the PAA mechanism.  
Due to the fact that the encoder and decoder play different roles in shaping model performance, each component was progressively integrated into the network to isolate and quantify its specific impact on localization accuracy.

\begin{table}[t]
    \small
    \centering
    \caption{Impact of each core component. This table presents the results of ablation experiments conducted on key components of KAN Linear (KAN-L), PKAN, and PAA within the encoder and decoder. It evaluates the contribution of each component to the overall model performance, highlighting their individual impact on precision and efficiency. The best results are highlighted in \textbf{bold}.} 
    \vspace{0.1cm}
    \resizebox{0.7\columnwidth}{!}{
        \begin{tabular}{c|ccc|ccc|ccc}
        \toprule[1pt]
        \multirow{2}{*}{Base} & \multicolumn{3}{c|}{Encoder} & \multicolumn{3}{c|}{Decoder} & Prec. & \# Params. & FLOPs \\
        \cline{2-7}
           & KAN-L & PKAN & PAA & KAN-L & PKAN & PAA & (\%) & (M) & (G) \\ 
        \midrule
        \checkmark &  &  &  &  &  &  &39.1 & 47.42 &429.41 \\ 
        \checkmark & \checkmark &  &  &  &  &  &43.6  & 42.11 & 407.26 \\
        \checkmark & \checkmark  &\checkmark &  &  &  &  & 47.9 & 42.10 & 406.26 \\
        \checkmark & \checkmark & \checkmark & \checkmark &  &  &  & 50.7 &46.52 & 396.35 \\
        \checkmark & \checkmark & \checkmark & \checkmark & \checkmark &  &  & 60.9 & 41.21& 392.56 \\ 
        \rowcolor{gray!20}
        \checkmark & \checkmark & \checkmark &\checkmark  & \checkmark & \checkmark &  &\textbf{65.9} & 41.21 & 391.57 \\
        \checkmark & \checkmark &\checkmark  & \checkmark & \checkmark &\checkmark  & \checkmark &59.9 &45.01 & 379.12 \\
        \bottomrule[0.1pt]
    \end{tabular}
  }
 \label{tab:components}
\end{table}

The ablation study yields several noteworthy findings. First, incorporating KAN-based linear layers leads to a substantial reduction in model parameters without sacrificing accuracy. In contrast to conventional linear layers with fixed weights, KAN layers adaptively adjust their transformations based on the input distribution, providing a more flexible and efficient alternative. Second, although the proposed PKAN modules do not significantly decrease the parameter count compared to standard MLPs, they markedly improve the network representational capacity, resulting in improved localization accuracy. Third, replacing the self-attention mechanism with the proposed PAA module further improves performance. Although this substitution introduces a slight increase in parameters, it yields clear gains in precision and reduces computational cost by employing additive operations to more efficiently capture sparse spatial dependencies.
In summary, the modular design of the AKT framework, integrating KAN linear layers, PKAN modules, and the PAA mechanism, achieves a favorable balance between computational efficiency and representational power, thus ensuring robust and scalable performance for maize core localization in real-world agricultural remote sensing scenarios.

Incorporating novel kernel functions into the PKAN module enables a more effective extraction of informative features from complex input data, thereby improving the accuracy of the localization. This capability is particularly beneficial in high-resolution UAV imagery, where high-dimensional inputs and non-linear feature distributions are common. Furthermore, the proposed PAA mechanism demonstrates its effectiveness as an alternative to conventional self-attention by efficiently capturing contextual information within Transformer-based architectures.  

However, it should be noted that while the three core components integrated into the encoder consistently improved performance, the introduction of PAA into the decoder led to a decrease in accuracy. Taking into account both the experimental results and the complexity of the model, PAA was therefore excluded from the decoder design. A closer analysis suggests that directly incorporating PAA in the decoder may introduce redundant or conflicting contextual information, thereby leading to suboptimal decoding performance. Addressing this limitation will be an important direction for future work.

\subsubsection{Impact of Padé KAN Module}
\label{abl_PKAN}
To comprehensively evaluate the efficacy of KAN within the proposed AKT framework, this section presents systematic comparative analyzes that examine the linear layer of KAN in multiple dimensions and various variants of KAN~\cite{fastkan, effkan}. As demonstrated in Table~\ref{tab:kan}, the experimental results indicate that the dimensionality of the KAN linear layer of 32 yields optimal performance. This configuration establishes an effective balance between computational efficiency and localization accuracy, representing the most suitable parameterization for the AKT architecture in agricultural remote sensing applications.
Furthermore, extensive comparative evaluations between conventional MLP architectures and various KAN implementations were conducted through systematic ablation studies within the framework. Quantitative results demonstrate that PKAN consistently outperforms alternative KAN variants, achieving superior localization accuracy while maintaining reduced parameter complexity and computational overhead, such as parameters and FLOPs.
This computational efficiency renders PKAN particularly advantageous for deployment in resource-constrained scenarios characteristic of real-time, high-resolution agricultural monitoring systems.
These empirical findings substantiate the effectiveness of PKAN in improving both the representational capacity and adaptive capabilities of the AKT framework, ensuring robust performance for precision maize core localization across diverse agricultural environments and imaging conditions.

\begin{table}[h]
\small
\centering
\caption{Impact of KAN components. Evaluation of different hidden sizes (dim) for the KAN linear layer, along with comparisons of MLP and various KAN variants used as replacements for PKAN. The best results are highlighted in \textbf{bold}.}
\vspace{0.1cm}
\resizebox{0.9\columnwidth}{!}{
\begin{tabular}{ccccc|cccc}
\midrule
Dim &Prec.(\%)&Param.(M) &FLOPs(G) &&Variant  &Prec.(\%)  &Param.(M) & FLOPs(G) \\
\midrule
8   &31.8  &39.73  &389.38  && MLP   & 60.9 & 41.21 & 392.56    \\
16  &  32.2   &40.22 &390.44  && KAN  & 52.71  & 52.4 & 522.12     \\
32  & \textbf{65.9} &41.2  &392.57 && Efficient KAN    &50.3   & 84.56 & 907.96    \\ 
64   &  31.5    &43.2  &396.81  && Fast KAN    & 47.1   & 57.97 & 734.30   \\
128  &  32.5   &47.11  &405.30   && PKAN     & \textbf{65.9}  & 41.21 & 392.57   \\ 
\midrule
\end{tabular}
}
\label{tab:kan}
\end{table}

\subsubsection{Impact of PKAN Additive Attention Module}
The contribution of the proposed PKAN Additive Attention (PAA) module was quantitatively assessed through systematic ablation studies, with comprehensive results presented in Table~\ref{tab:PAA}. The removal of the PAA module resulted in a substantial degradation in localization performance, with precision metrics falling from 65.9\% to 47.9\%. This marked reduction of 18.0\% underscores the critical importance of the PAA module in achieving accurate maize localization in complex agricultural scenes.
This performance differential is attributed to the inherent limitations of conventional self-attention mechanisms in modeling the complex spatial dependencies and heterogeneous feature distributions characteristic of agricultural remote sensing imagery. The PAA module addresses these challenges by efficiently encoding high-dimensional feature representations while maintaining computational tractability, thereby achieving superior localization accuracy with reduced computational overhead.
To further validate the architectural advantages of the proposed framework, comprehensive comparative analyses were conducted by systematically replacing PKAN with alternative architectures, including standard MLP, vanilla KAN, Efficient KAN~\cite{efficienKAN}, and Fast KAN~\cite{fKAN}.
These comparative implementations exhibited substantial variations in parameter complexity and failed to achieve an optimal trade-off between computational efficiency and localization precision. 
In particular, the vanilla KAN architecture demonstrated limited adaptability to complex and cluttered backgrounds and varying illumination conditions prevalent in operational agricultural monitoring scenarios, resulting in degraded localization performance under real-world deployment conditions.

The architectural sensitivity to attention module sequencing was systematically investigated through controlled permutation experiments. Specifically, the impact of reversing the spatial and channel attention module order within the PAA framework was evaluated. Although this architectural modification produced negligible variations in model complexity metrics, including parameters, FLOPs, and FPS, it resulted in measurable degradation of localization performance. 
Based on these empirical findings, the configuration $CA(SA(\cdot))$, in which channel attention follows spatial attention, was adopted as the optimal architecture to extract hierarchical global characteristics from high-resolution agricultural imagery.

The optimized configuration demonstrates superior performance, achieving a localization precision of 65.9\% and a Mean Absolute Error (MAE) of 7.1 pixels, thus validating the effectiveness of the proposed PAA module for precision agricultural applications. These quantitative results confirm that the sequential application of spatial and channel attention mechanisms provides enhanced feature representation capabilities essential for accurate maize location in complex field environments.

\begin{table}[t]
    \centering
    \setlength{\abovecaptionskip}{0pt}
    \caption{Impact of the PAA module. The ablation experiments conducted to evaluate the impact of the PAA module are summarized by replacing PKAN and modifying the architecture. The best results are highlighted in \textbf{bold}.} 
    \vspace{0.1cm}
    \label{tab:PAA}
    \resizebox{0.65\columnwidth}{!}{
        \begin{tabular}{lccccc}
            \toprule
            Method & Prec.(\%) &MAE &Params.(M) & FLOPs(G) & FPS \\ 
            \midrule  
            Self-attention  &47.9 &7.6 &36.8  & 403.47   &21.59 \\
            Linear   & 55.7    &7.9    &38.5  & 381.29  &21.21 \\
            w/ MLP   & 59.1  &7.3    &41.2  &392.56    &21.19  \\
            w/ KAN   & 55.2  &7.6    &42.0  & 394.18   &19.24  \\
            \midrule
            SA(CA($\cdot$))  &63.9 &7.2   &41.2  &394.57  &22.45  \\
            PAA     & \textbf{65.9}   & \textbf{7.1}  & 41.2 & 392.57  & \textbf{23.05}  \\
            \bottomrule
        \end{tabular}
    }
\end{table}

\section{Discussion}
\label{sec:discussion}
To comprehensively evaluate the generalizability and practical applicability of AKT, the experimental validation is extended to encompass additional remote sensing tasks critical for precision agriculture. Specifically, Section \ref{countingper} presents quantitative assessments of performance in crop counting applications, while Section \ref{placingper} evaluates efficacy in estimating plant spacing for field monitoring. Furthermore, Section \ref{limitation} provides a critical analysis of methodological constraints and identifies potential avenues for future research. These extended evaluations demonstrate the versatility of the proposed approach beyond single-point localization tasks, establishing broader utility for comprehensive agricultural remote sensing applications.

\subsection{Counting Performance}
\label{countingper}
This section presents a comprehensive quantitative evaluation of the counting performance of the proposed method through a systematic comparison with state-of-the-art approaches. As demonstrated in Table~\ref{tab:count}, the proposed AKT framework consistently achieves superior or highly competitive performance in all experimental configurations.
\textit{Evaluation Metrics.} Following established protocols in the literature on density estimation and object counting~\cite{p2pnet, tasselnetv3}, Mean Absolute Error (MAE) and Mean Squared Error (MSE) are adopted as the primary evaluation metrics. MAE quantifies the average absolute deviation between the predicted counts and GT annotations, providing a robust measure of counting accuracy that is invariant to the direction of the error. MSE, through its quadratic penalty function, assigns greater weight to large prediction errors, thereby offering enhanced sensitivity to significant counting discrepancies and spatial distribution anomalies. The complementary nature of these metrics enables a comprehensive performance assessment, effectively balancing overall accuracy with robustness to outlier predictions, critical considerations for operational agricultural monitoring applications.

The proposed approach leverages high-resolution feature representations to extract discriminative spatial patterns, allowing the precise localization of objects in complex agricultural scenes. Quantitative evaluation of the test dataset reveals that our method achieves optimal performance with MAE of 7.1 and MSE of 25.4, outperforming existing localization-based counting methods in both validation and test datasets. These results demonstrate the efficacy of the proposed architecture in maintaining spatial precision while achieving accurate count estimates.
The observed disparity between MAE and MSE metrics warrants further analysis. The relatively low MAE indicates robust performance across the majority of samples, suggesting reliable count estimation under standard operating conditions. In contrast, the elevated MSE value reveals the presence of occasional large prediction errors in challenging scenarios, where squared deviations disproportionately impact the metric. This phenomenon typically occurs in complex agricultural scenes characterized by severe occlusions, irregular plant distributions, or extreme illumination variations. A comprehensive analysis of these failure modes and their implications for practical deployment is presented in Section~\ref{limitation}.

\begin{table}[h]
\centering
\setlength{\tabcolsep}{4.1mm}
\caption{Counting performance of various methods on the maize core localization validation and test datasets. The best results are highlighted in \textbf{bold}, $\downarrow$ means that a smaller value is better.}
\vspace{0.1cm}
\label{tab:count}
\resizebox{0.65\columnwidth}{!}{
\begin{tabular}{ccccc}
\toprule
{\multirow{2}{*}{Method}} &\multicolumn{2}{c}{Validation set} &\multicolumn{2}{c}{Test set} \\ 
\cmidrule{2-5}
&MAE $\downarrow$ &MSE $\downarrow$ &MAE $\downarrow$ &MSE $\downarrow$ \\
\midrule
LCFCN\cite{LCFCN} &11.8 &59.8 &9.6 &54.9 \\
LSC-CNN\cite{LSCCNN} &10.8 &58.1 &9.5 &54.0 \\
GPR\cite{GPR} &9.7 &54.4 &9.4 &52.4 \\
SDNet\cite{SDNet} &9.2 &48.8 &9.0 &46.2 \\
SCAR\cite{gao2019scar} &9.8 &55.9 &9.5 &53.2 \\
RAZ\_loc\cite{liu2019recurrent} &8.8 &44.8 &8.7 &42.4 \\
AutoScale\_loc\cite{xu2019autoscale} &8.7 &40.1 &8.4&38.1 \\
GL\cite{GL} &8.3 &34.1 &8.0 & 31.9 \\
SCALNet\cite{SCALNet} &9.4 &50.7 &9.1 &47.9 \\
TopoCount\cite{topo} &8.6 &37.9 &8.2 &35.6 \\
FIDTM\cite{FIDTM} &8.9 &44.7 &8.6 &42.0\\
CLTR\cite{CLTR} &8.1 & 32.9 &7.91 & 32.8 \\
AKT (Ours) &\textbf{7.3} &\textbf{29.1}&\textbf{7.1}&\textbf{25.4} \\
\bottomrule
\end{tabular}}
\end{table}

\subsection{Plant Spacing Analysis}
\label{placingper}
Plant spacing represents a fundamental biophysical parameter in precision agriculture, directly governing crop yield potential, resource utilization efficiency, and site-specific management strategies~\cite{biwen}. Accurate quantification of the distances between plants from high-resolution UAV imagery facilitates automated assessment of the uniformity of the planting, the quality of the establishment of the stand and the optimal management zone delineation, critical components for sustainable agricultural intensification~\cite{UAV2}.

Therefore, in this section, the geometric measurement capabilities of the proposed AKT framework for quantitative agricultural remote sensing applications are evaluated. The methodology employs precise localization output to compute interplant spacing through Euclidean distance calculations between adjacent maize tassel centroids in the image coordinate system. Following established remote sensing validation protocols, ground truth measurements were acquired through systematic field surveys using calibrated measuring equipment. Specifically, manual measurements of the distances between plants were collected for 59 maize plants distributed in multiple experimental plots, providing independent reference data for a rigorous evaluation of geometric accuracy.
The validation framework adheres to standard remote sensing accuracy assessment methodologies, in which image-derived geometric measurements are systematically compared against in situ observations to quantify measurement uncertainty and establish confidence intervals for operational deployment. This comprehensive validation approach ensures the reliability of the proposed method for practical agricultural monitoring applications where spatial precision is paramount.

\begin{figure}[t]
	\centering
	\includegraphics[width=1\linewidth]{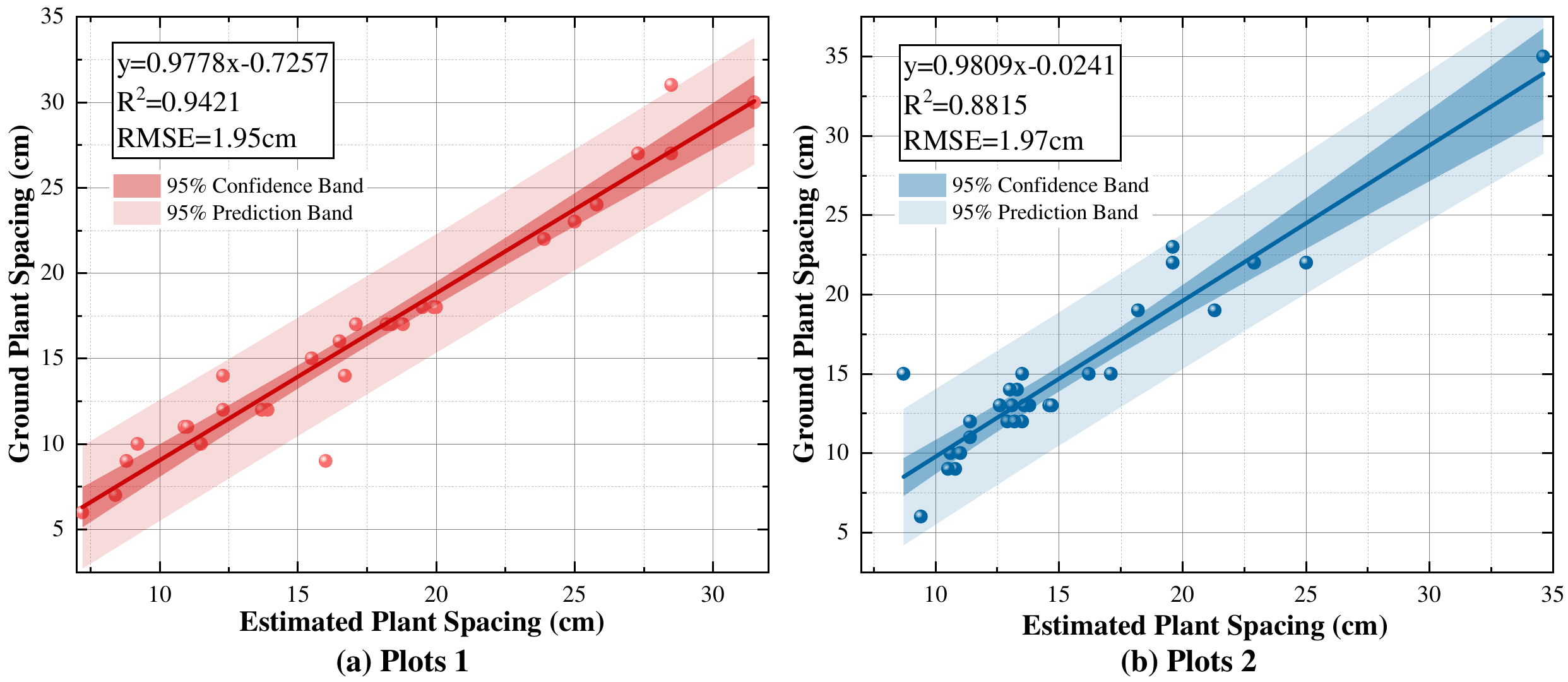}
	\caption{An unseen plot was used to evaluate the performance of the localization method in estimating plant distances.}
	\label{fig:spacing}
\end{figure}

\begin{figure}[t]
	\centering
	\includegraphics[width=1\linewidth]{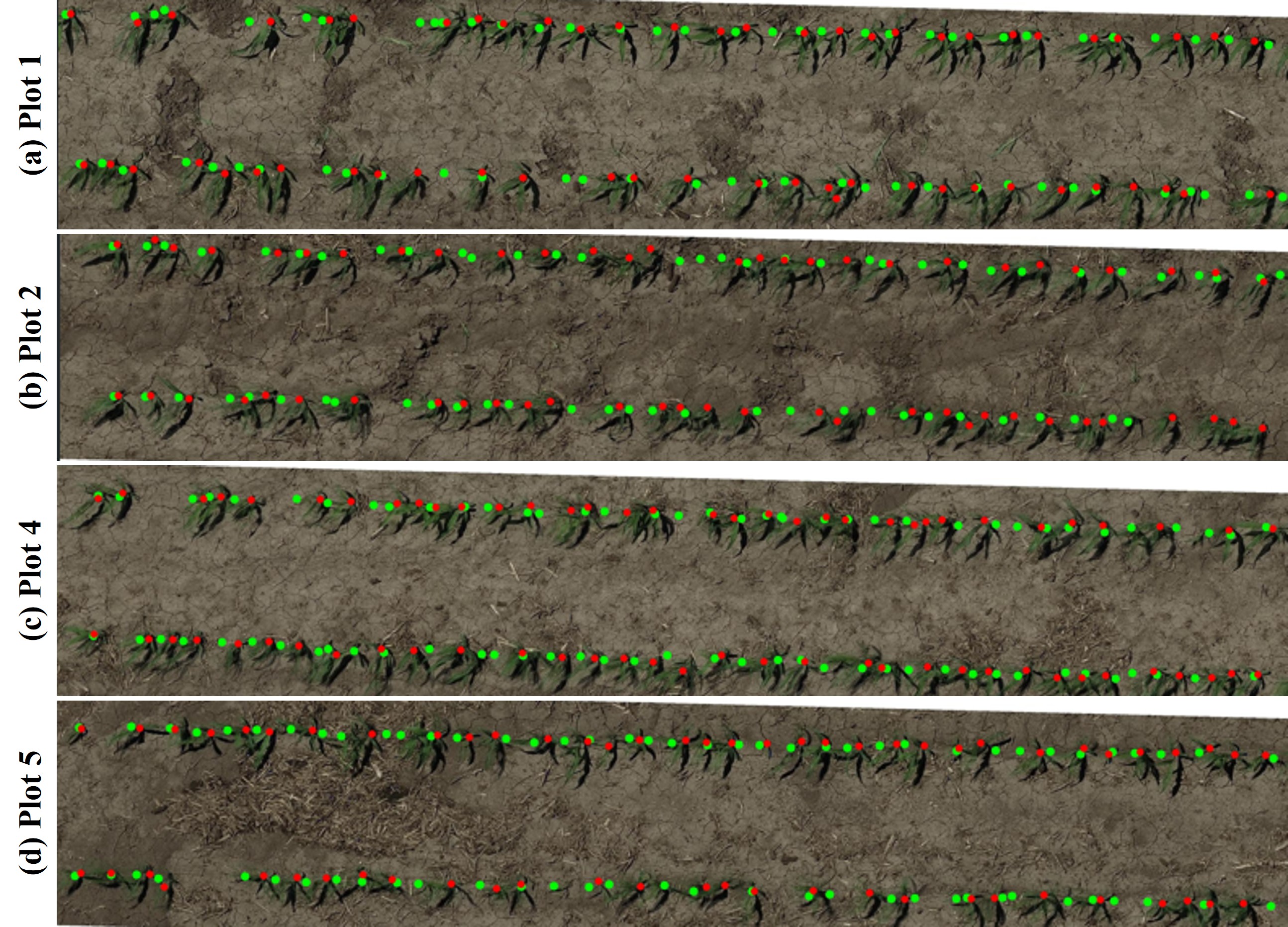}
	\caption{Qualitative visualizations of the proposed AKT methods on unseen plot samples.}
	\label{fig:plotvis}
\end{figure}

\textit{Geometric Accuracy Assessment.} The geometric fidelity of UAV-derived plant spacing measurements was quantitatively evaluated using established remote sensing validation protocols. The distance between plants \((PS_{dist})\) was calculated as the Euclidean distance between the centroids of adjacent plants in the image coordinate system~\cite{UAV2}.

\begin{equation}
PS_{dist} =\sqrt{\left ( ex_{n-1} - ex_{n} \right ) ^{2} + \left ( ey_{n-1} - ey_{n} \right ) ^{2} }
\label{spacing}
\end{equation}
where \(ex\) and \(ey\) denote the predicted centroid coordinates in the pixel space, and \(n\) represents the sequential plant identifier.

Root Mean Square Error (RMSE) served as the primary precision metric, quantifying the geometric discrepancy between UAV-derived measurements and in situ reference data. RMSE values are reported in absolute units (centimeters) to facilitate direct comparison with conventional survey methods and evaluate operational viability for precision agriculture applications.
The visual validation results for independent test plots are presented in Fig.\ref{fig:plotvis}, while Fig.\ref{fig:spacing} illustrates the quantitative comparison between plant spacing derived from remote sensing and field measured using Eq.~\ref{spacing}.
Geometric validation revealed strong agreement between UAV-derived measurements and GT observations. Linear regression analysis yielded determination coefficients \((R^{2})\) of 0.9421 and 0.8815 for the respective test plots, indicating that the remote sensing approach captures over 88\% of the variance in the actual spacing of the plant. This level of precision satisfies the operational requirements for agricultural monitoring applications~\cite{UAV1,UAV2}.

The proposed AKT framework achieved RMSE values of 1.95--1.97 cm, demonstrating sub-2 cm accuracy comparable to the 1.89 cm reported previously~\cite{UAV1}. Notably, our method maintains this precision while processing full-resolution imagery of \(3648 \times 4864\) pixels, representing a 265-fold increase in spatial coverage compared to the \(224 \times 224\) pixel patches required by existing methods.
The achieved geometric precision validates the framework's capability for operational deployment in precision agriculture applications requiring plant-level measurements.

These findings establish the viability of the proposed approach for automated extraction of critical biophysical parameters from UAV imagery. The demonstrated geometric precision supports various agricultural applications including stand uniformity assessment, emergence monitoring, and management zone delineation. The framework's capacity to deliver reliable plant-level measurements from high-resolution aerial imagery presents substantial advantages over conventional ground-based methods in terms of spatial coverage, temporal efficiency, and seamless integration with digital agriculture systems.

\subsection{Limitations and Future Directions}
\label{limitation}
\begin{figure}[t]
	\begin{center}
		\includegraphics[width=1\linewidth]{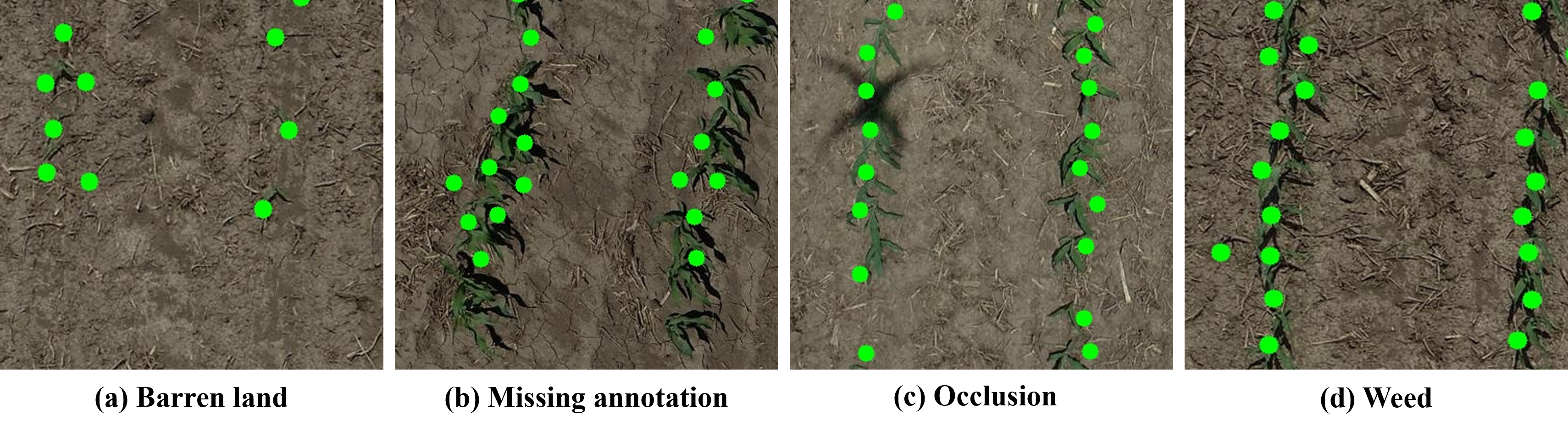}
	\end{center}
    \vspace{-0.5 cm}
	\caption{Some failed samples and challenging scenarios are presented. From left to right, these factors include barren land (sample sparsity), missing annotations, occlusions, and weeds, among others. These challenges hindered the performance of the model in real-world maize core localization.}
	\label{fig:challenge}
\end{figure}

This study addresses fundamental challenges in the location of the maize core in high-resolution UAV imagery, with representative failure cases illustrated in Fig.~\ref{fig:challenge}. Despite the demonstrated effectiveness of the proposed framework, several limitations warrant discussion to guide future research directions.

\textbf{Dataset Constraints.} The availability of high-quality annotated datasets for agricultural remote sensing applications remains a critical bottleneck for supervised learning approaches. While this study contributes a carefully curated dataset with precise point annotations, the scale remains limited by the substantial resources required for expert annotation. Future research should prioritize the development of semi-supervised and weakly-supervised learning strategies to leverage unlabeled UAV imagery, alongside active learning approaches to optimize annotation efficiency.

\textbf{Environmental Variability.} Complex field conditions present persistent challenges for robust localization. Bare soil regions (Fig.\ref{fig:challenge}(a)) introduce spatial ambiguity, while canopy overlap in dense plantings (Fig.\ref{fig:challenge}(c)) causes significant occlusion of individual plants. Irregular planting patterns lead to spatial discontinuities in the annotation space (Fig.\ref{fig:challenge}(b)), and weed interference (Fig.\ref{fig:challenge}(d)) generates false positive detections. In addition, adverse atmospheric conditions, including precipitation and fog, degrade image quality and discriminability of characteristics. Future work should explore multi-modal fusion strategies incorporating multispectral and thermal imagery to enhance robustness under challenging environmental conditions.

\textbf{Computational Efficiency.} Real-time processing requirements for operational precision agriculture applications require further architectural optimization. Although the proposed PKAN and PAA modules reduce computational complexity compared to standard Transformer architectures, deployment on resource-constrained UAV platforms remains challenging. Future research should investigate knowledge distillation and neural architecture search techniques to develop lightweight variants suitable for edge computing scenarios.

\textbf{Methodological Considerations.} The current point matching strategy employing the \(\mathcal{L}_{1}\) distance exhibits inherent ambiguities in dense object scenarios, occasionally resulting in suboptimal assignments. The development of learnable matching strategies that incorporate spatial context and geometric constraints represents a promising direction to improve the precision of localization in cluttered agricultural scenes.

\section{Conclusion}
\label{sec:conclusion}
This study presents a comprehensive framework for precision maize localization in high-resolution UAV imagery, addressing critical challenges in agricultural remote sensing applications. A novel Point-annotated Maize Localization dataset (PML) is introduced, comprising varied field conditions essential for developing robust agricultural monitoring systems. The Additive Kolmogorov-Arnold Transformer (AKT) framework leverages mathematical foundations from the Kolmogorov-Arnold representation theorem to achieve efficient feature decomposition while maintaining computational tractability for operational deployment.
The AKT architecture demonstrates superior performance across multiple evaluation metrics, achieving a localization precision of 65.9\% and MAE of 7.1 in the PML. Comprehensive ablation studies validate the effectiveness of the proposed PKAN and PAA modules in enhancing representational capacity while reducing computational overhead compared to standard Transformer architectures. The framework maintains robust performance under challenging field conditions, including occlusion of the plant, irregular spacing, and varying stages of growth.
Quantitative comparisons with state-of-the-art methods reveal that AKT achieves optimal trade-offs between localization accuracy and computational efficiency, with inference speeds suitable for real-time agricultural monitoring applications. Furthermore, the framework demonstrates strong generalization capabilities, achieving sub-2 cm RMSE in plant spacing estimation tasks, comparable to ground-based measurements while processing full-resolution imagery of \(3648 \times 4864\) pixels.

The demonstrated effectiveness of integrating mathematical principles with deep learning architectures establishes a promising direction for agricultural remote sensing research. The proposed methodology extends beyond the localization of maize, offering scalable solutions for various precision agriculture applications including stand count evaluation, phenotypic analysis, and yield prediction.
Future research will investigate multi-modal data fusion strategies, develop lightweight variants for edge deployment, and extend the framework to additional crop species. These advances will contribute to sustainable agricultural intensification through enhanced monitoring capabilities, supporting global food security objectives through data-driven precision management strategies.

\bibliography{refs}

\end{document}